\documentclass[11pt,a4paper]{article}
\usepackage[utf8]{inputenc}
\usepackage[margin=1in]{geometry}
\usepackage{amsmath,amssymb}
\usepackage{graphicx}
\usepackage{booktabs}
\usepackage{multirow}
\usepackage{array}
\usepackage{xcolor}
\usepackage[hyperfootnotes=false]{hyperref}
\usepackage{enumitem}
\usepackage[numbers]{natbib}
\usepackage{times}
\usepackage[section]{placeins}

\title{\textbf{RadLite}: Multi-Task LoRA Fine-Tuning of Small Language Models\\for CPU-Deployable Radiology AI}

\author{
  Pankaj Gupta, MD$^{1,*}$ \and Kartik Bose$^{1}$\\[6pt]
  $^{1}$Postgraduate Institute of Medical Education and Research, Chandigarh, India\\[4pt]
  \small{$^{*}$Corresponding author: Additional Professor, Department of Radiodiagnosis and Imaging,}\\
  \small{Postgraduate Institute of Medical Education and Research, Chandigarh, India}
}

\date{}

\begin{document}
\maketitle

\begin{abstract}
Large language models (LLMs) show promise in radiology but their deployment is limited by computational requirements that preclude use in resource-constrained clinical environments.
We investigate whether small language models (SLMs) of 3--4 billion parameters can achieve strong multi-task radiology performance through LoRA fine-tuning, enabling deployment on consumer-grade CPUs.
We train \textbf{Qwen2.5-3B-Instruct} and \textbf{Qwen3-4B} on 162K samples spanning 9 radiology tasks - RADS classification across 10 systems, impression generation, temporal comparison, radiology NLI, NER, abnormality detection, N/M staging, and radiology Q\&A - compiled from 12 public datasets.
Both models are evaluated on up to 500 held-out test samples per task with standardized metrics.
Our key findings are: (1) LoRA fine-tuning dramatically improves performance over zero-shot baselines (RADS accuracy +53\%, NLI +60\%, N-staging +89\%); (2) the two models exhibit complementary strengths - Qwen2.5 excels at structured generation tasks while Qwen3 dominates extractive tasks; (3) a task-routed oracle ensemble combining both models achieves the best performance across all tasks; (4) few-shot prompting with fine-tuned models \emph{hurts} performance, demonstrating that LoRA adaptation is more effective than in-context learning for specialized domains; and (5) models can be quantized to GGUF format ($\sim$1.8--2.4GB) for CPU deployment at 4--8 tokens/second on consumer hardware.
Our work demonstrates that small, efficiently fine-tuned models - which we collectively call \textbf{RadLite} - can serve as practical multi-task radiology AI assistants deployable entirely on consumer hardware without GPU requirements.
Code and models are available at \url{https://github.com/RadioX-Labs/RadLite}.
\end{abstract}

\section{Introduction}

Radiology report interpretation requires understanding across multiple clinical contexts - from standardized classification systems (RADS) to temporal comparisons, anatomical entity recognition, and clinical reasoning through natural language inference.
While frontier proprietary models such as GPT-5.2 and Gemini~3 demonstrate strong performance on clinical NLP tasks~\cite{multirads2025,singhal2023large}, their deployment in clinical settings is constrained by cost, latency, data privacy concerns, and the need for internet connectivity~\cite{mesko2023impact}.
This is particularly problematic in resource-limited healthcare settings, rural hospitals, and regions with limited internet infrastructure, where on-premise AI tools could have the greatest impact.

Recent advances in open-weight small language models (SLMs) offer a promising alternative.
Models in the 3--4 billion parameter range, such as Qwen2.5-3B-Instruct~\cite{qwen2025} and Qwen3-4B~\cite{qwen3}, achieve impressive general-purpose performance, raising the question: \emph{can these models be efficiently adapted to serve as comprehensive radiology AI assistants that run entirely on consumer hardware?}

In this work, we present \textbf{RadLite}, a systematic study of multi-task LoRA fine-tuning~\cite{hu2022lora} for radiology applications targeting CPU deployment.
Our contributions are:

\begin{enumerate}[leftmargin=*]
  \item A comprehensive benchmark evaluating two SLMs across 9 radiology tasks compiled from 12 public datasets, including RADS classification across 10 systems and 8 additional clinical NLP tasks.
  \item Detailed zero-shot vs.\ fine-tuned comparison demonstrating the dramatic impact of LoRA adaptation across all tasks.
  \item Analysis of complementary strengths between Qwen2.5-3B and Qwen3-4B, with an oracle ensemble (an upper-bound combination that routes each task to the better-performing model) achieving best-in-class performance.
  \item Empirical evidence that few-shot prompting degrades fine-tuned model performance, supporting the superiority of parameter-efficient fine-tuning over in-context learning for domain specialization.
  \item Demonstration that the resulting models can be quantized to $\sim$1.8--2.4GB and deployed on consumer CPUs at 4--8 tokens/second, eliminating the need for GPU infrastructure.
\end{enumerate}

\section{Related Work}

\textbf{LLMs in Radiology.}
The application of large language models to radiology has grown rapidly.
Frontier models have been evaluated on tasks including report generation~\cite{bannur2023temporal}, impression summarization~\cite{zhang2018summarize}, and clinical NLP benchmarks~\cite{singhal2023large}.
RadGraph~\cite{jain2021radgraph} introduced a structured NER schema for radiology entities and relations, subsequently extended by RadGraph2~\cite{delbrouck2024radgraph2} with temporal change annotations.
The CheXbert labeler~\cite{chexbert2020} automated multi-label abnormality classification from chest X-ray reports.
The Multi-RADS benchmark~\cite{multirads2025} evaluated 41 open-weight SLMs (0.135--32B) alongside GPT-5.2 across 10 RADS systems, finding that models below 1B achieved only 27\% accuracy, the 1--10B range averaged 57.5\%, and GPT-5.2 reached 81.1\% under guided prompting - establishing that RADS classification scales strongly with model capacity and motivating task-specific fine-tuning of small models.
Several works have explored LLMs for TNM staging and abnormality detection~\cite{chexbert2020}.
However, all prior work focuses on either single-task evaluation or requires models too large for local deployment.
Our work addresses this gap by training specialized small models that outperform zero-shot baselines on the same tasks while being deployable on consumer hardware.

\textbf{Parameter-Efficient Fine-Tuning.}
LoRA~\cite{hu2022lora} enables efficient adaptation by training low-rank decomposition matrices alongside frozen weights, reducing trainable parameters by over 98\%.
QLoRA~\cite{dettmers2024qlora} further combined quantization with LoRA for memory-efficient training.
These methods have been applied to medical domain adaptation~\cite{han2023medalpaca,chen2023meditron}, but typically for single-task scenarios or using models of 7B+ parameters.
Multi-task LoRA training remains underexplored in the medical domain, particularly for small models where training capacity is limited.
Recent work on LoRA composition~\cite{huang2024lorahub} has shown promise for combining task-specific adapters, but requires separate training per task.
We demonstrate that a single LoRA adapter trained on 9 diverse tasks simultaneously can achieve strong performance across all tasks.

\textbf{Small Language Models in Healthcare.}
The trend toward smaller, deployable models is growing~\cite{thirunavukarasu2023}, motivated by privacy, cost, and latency requirements.
Models such as Phi-3~\cite{abdin2024phi3}, Gemma~\cite{gemma2024}, and the Qwen family~\cite{qwen2025,qwen3} demonstrate that 3--4B parameter models can achieve competitive performance with proper training.
In healthcare specifically, domain-specific SLMs like Meditron-70B~\cite{chen2023meditron} and BioMistral~\cite{labrak2024biomistral} have shown that medical knowledge can be instilled through continued pre-training.
However, these models still require GPU inference and have not been optimized for multi-task clinical NLP on CPU.
Our work shows that with LoRA fine-tuning, even smaller models (3--4B) can achieve strong multi-task radiology performance and be deployed via quantization on consumer CPUs.

\textbf{Model Quantization and CPU Deployment.}
Post-training quantization techniques such as GPTQ~\cite{frantar2023gptq}, AWQ~\cite{lin2024awq}, and GGUF-format quantization~\cite{ggml2023} enable dramatic model compression with minimal accuracy loss.
The GGUF format supports 4-bit quantization (Q4\_K\_M) that reduces model size by 3--4$\times$ while maintaining inference quality.
Combined with CPU inference engines like llama.cpp, quantized models can run on standard consumer hardware without GPU support.
This is critical for clinical deployment in resource-limited settings, where dedicated GPU hardware may be unavailable.
To our knowledge, no prior work has demonstrated multi-task radiology AI deployment on consumer CPUs.

\section{Methods}

\subsection{Models}

We evaluate two open-weight instruction-tuned language models:

\begin{itemize}[leftmargin=*]
  \item \textbf{Qwen2.5-3B-Instruct}~\cite{qwen2025}: A 3-billion parameter model with 36 transformer layers, hidden dimension 2048, 16 attention heads, and 8 key-value groups (GQA).
  \item \textbf{Qwen3-4B}~\cite{qwen3}: A 4-billion parameter model featuring a hybrid thinking architecture that supports both reasoning and direct response modes.
\end{itemize}

Both models use Grouped Query Attention (GQA) and SwiGLU activation functions.
Qwen3 additionally supports a thinking mode where the model generates internal reasoning before producing outputs.
We select these models based on their strong performance on the Multi-RADS zero-shot benchmark~\cite{multirads2025}, where both Qwen2.5 and Qwen3 variants ranked among the top-performing open-weight families in the 1--10B class, and to investigate whether their architectural differences affect multi-task radiology performance.

\subsection{Training Data}

Our training corpus consists of 161,586 samples spanning 9 radiology tasks compiled from 12 publicly available datasets.
Below we describe each task and its data sources; Table~\ref{tab:datasets} provides a comprehensive summary.

\begin{itemize}[leftmargin=*]
  \item \textbf{Impression Generation} (30,000 samples): Generating concise impression sections from radiology report findings.
  Data sourced from MIMIC-CXR~\cite{johnson2019mimiccxr} (chest X-ray), MR-RATE (brain/spine MRI), and CT-RATE~\cite{hamamci2024ctrate} (body CT).
  Reports were split into findings and impression sections; the task prompt presents the findings and requests the impression.

  \item \textbf{Radiology Q\&A} (30,000 samples): Answering clinical questions about radiology images and reports.
  Data sourced from Radialog-Instruct~\cite{radialog2024} (open-domain radiology QA from PMC literature), LLaVA-Med~\cite{li2023llavamed} (biomedical QA text pairs), and MedQA-USMLE~\cite{jin2021medqa} (USMLE-style questions with radiology content).

  \item \textbf{Temporal Comparison} (30,000 samples): Identifying changes (new, worsened, improved, resolved, unchanged) between findings across serial radiology reports.
  Data sourced from RadGraph2~\cite{delbrouck2024radgraph2} and RadGraph2-inference, which annotate temporal changes in MIMIC-CXR reports over time.

  \item \textbf{Radiology NER} (25,000 samples): Extracting anatomical structures, observations (present/absent/uncertain), and change indicators from reports using the RadGraph schema.
  Data sourced from RadGraph2~\cite{delbrouck2024radgraph2}, RadGraph2-inference, RadGraph-XL~\cite{miyanishi2023radgraph}, and PIFIR-PET-CT.

  \item \textbf{N-staging} (19,554 samples): Predicting nodal staging (N0, N1, N2) from body CT reports.
  Data sourced from the Merlin dataset~\cite{blankemeier2026merlin}, a large-scale CT vision-language dataset; reports were screened for explicit lymph node descriptors to extract staging labels.

  \item \textbf{Abnormality Detection} (14,962 samples): Multi-label classification of chest X-ray abnormalities from the MIMIC-CXR dataset labeled by CheXbert~\cite{chexbert2020} (14 conditions: consolidation, edema, enlarged cardiomediastinum, fracture, lung lesion, etc.).
  Reports converted to instruction format asking to classify each finding as positive/negative/uncertain.

  \item \textbf{RADS Assignment} (9,355 samples): Classification across 10 RADS systems - BI-RADS, PI-RADS, LI-RADS (including LR-TR treatment response), TI-RADS, CAD-RADS, VI-RADS, Lung-RADS, O-RADS, NI-RADS, and GB-RADS.
  Data curated from PubMed Central literature reports, LLM-assigned labels, synthetic reports covering under-represented categories, and expert-collected clinical reports~\cite{multirads2025}.
  Initial corpus of 9,520 samples was cleaned to 9,355 through deduplication and quality filtering.

  \item \textbf{M-staging} (2,235 samples): Predicting metastatic staging (M0, M1) from body CT reports.
  Data sourced from the Merlin dataset~\cite{blankemeier2026merlin}; reports were screened for explicit distant metastasis descriptors to extract staging labels.

  \item \textbf{Radiology NLI} (480 samples): Natural language inference on radiology report sentence pairs (entailment, contradiction, neutral).
  Data sourced from RADNLI~\cite{miura2021radnli}, a radiology-specific NLI benchmark.
\end{itemize}

\begin{table*}[htbp]
\centering
\resizebox{\textwidth}{!}{%
\begin{tabular}{l r l l l}
\toprule
\textbf{Task} & \textbf{Samples} & \textbf{Source Dataset(s)} & \textbf{Modality} & \textbf{Curation} \\
\midrule
Impression Gen. & 30,000 & MIMIC-CXR, MR-RATE, CT-RATE & CXR, MRI, CT & Findings--impression pairs \\
Radiology Q\&A & 30,000 & Radialog-Instruct, LLaVA-Med, MedQA & Multi & Question--answer pairs \\
Temporal Comp. & 30,000 & RadGraph2 / RadGraph2-inference & CXR & Temporal annotation \\
Radiology NER & 25,000 & RadGraph2, RadGraph-XL, PIFIR-PET-CT & CXR, CT, PET & Entity annotation \\
N-staging & 19,554 & Merlin (screened) & CT & LN descriptor screening \\
Abnormality Det. & 14,962 & MIMIC-CXR-CheXbert, RadGraph2 & CXR & Multi-label classification \\
RADS Assignment & 9,355 & Multi-RADS (PMC + synthetic + expert) & Multi & Multi-source curation$^\dagger$ \\
M-staging & 2,235 & Merlin (screened) & CT & Metastasis descriptor screening \\
Radiology NLI & 480 & RADNLI & CXR & Expert NLI annotation \\
\midrule
\textbf{Total} & \textbf{161,586} & \textbf{12 datasets} & & \\
\bottomrule
\end{tabular}%
}
\caption{Training dataset overview. Tasks, sample counts, source datasets, data modalities, and curation methods. High-volume tasks were capped at the listed amounts. $^\dagger$Includes LR-TR treatment response sub-category. CXR = Chest X-ray; CT = Computed Tomography; MRI = Magnetic Resonance Imaging; PET = Positron Emission Tomography; Multi = Multiple modalities.}
\label{tab:datasets}
\end{table*}

\begin{figure}[t]
\centering
\includegraphics[width=\columnwidth]{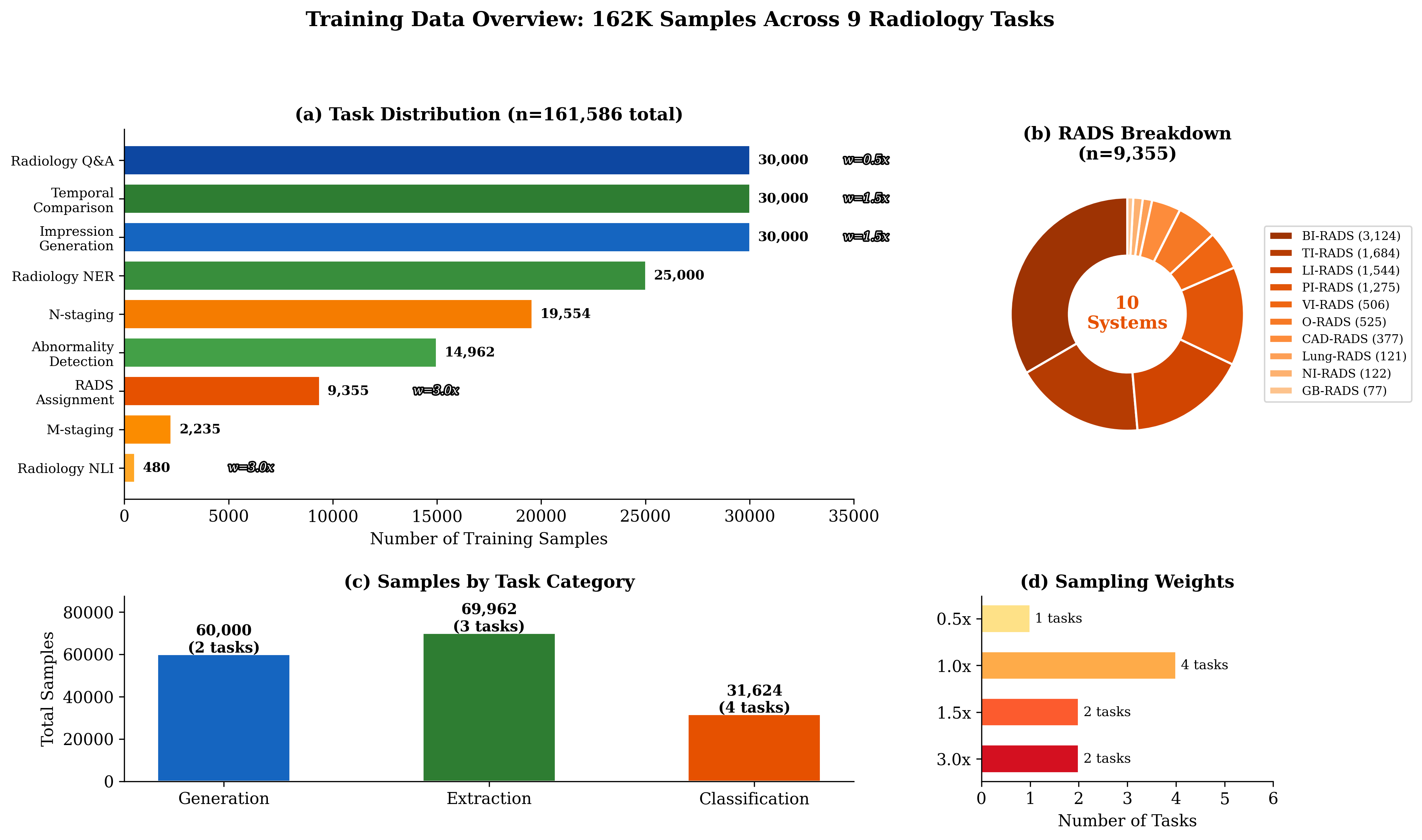}
\caption{Training data overview: 162K samples across 9 radiology tasks and 12 public datasets. (a) Task distribution by sample count with sampling weights. (b) RADS assignment breakdown across 10 classification systems. (c) Total samples by task category (Generation, Extraction, Classification). (d) Number of tasks per sampling weight tier.}
\label{fig:dataset}
\end{figure}

\textbf{Data Curation.}
All data was converted to a unified instruction-tuning format with task-specific prompts (e.g., \texttt{[TASK: rads\_assignment]} prefix) followed by the input text and expected output.
High-volume tasks (impression, temporal, NER, Q\&A) were capped at 30,000 samples through random subsampling to prevent task dominance during training.
Data quality was stratified into three tiers following source annotations: \emph{gold} (expert-annotated), \emph{silver} (model-predicted or algorithmically generated), and \emph{bronze} (LLM-assigned labels).
All samples underwent deduplication using exact-match filtering on input text and basic quality checks (non-empty output, valid label format).

\subsection{Task-Weighted Sampling}

To ensure balanced training across tasks with varying sample counts, we apply task-weighted sampling:

\begin{itemize}[leftmargin=*]
  \item \textbf{3$\times$ weight}: RADS assignment, Radiology NLI (critical clinical tasks with limited data)
  \item \textbf{1.5$\times$ weight}: Impression generation, Temporal comparison
  \item \textbf{0.5$\times$ weight}: Radiology Q\&A (high-volume, lower clinical priority)
  \item \textbf{1$\times$ weight}: All other tasks (NER, N-staging, M-staging, Abnormality detection)
\end{itemize}

This ensures that clinically critical tasks with small datasets (NLI: 480 samples, RADS: 9,355 samples) are sampled proportionally more often than high-volume tasks (Q\&A: 30,000 samples).

\subsection{LoRA Configuration}

Both models are fine-tuned using LoRA with identical configurations.
We selected LoRA rank $r=64$ and scaling factor $\alpha=128$ ($\alpha = 2r$, the standard recommended ratio~\cite{hu2022lora}) to provide sufficient adaptation capacity for 9 simultaneous tasks while keeping adapter sizes tractable ($\sim$240\,MB per model).
All seven projection matrices (\texttt{q\_proj}, \texttt{k\_proj}, \texttt{v\_proj}, \texttt{o\_proj}, \texttt{gate\_proj}, \texttt{up\_proj}, \texttt{down\_proj}) are targeted rather than a subset, ensuring that both attention and feed-forward sub-modules are adapted - an empirically important choice for tasks with diverse output formats spanning free-text generation and single-token classification.
Dropout of 0.05 provides light regularization appropriate for the large, diverse training corpus.
A learning rate of 2e-5 with cosine decay is used to avoid catastrophic forgetting of the base model's general language capabilities.
The effective batch size of 32 (8 per device $\times$ 4 gradient accumulation steps) balances GPU memory constraints with training stability.
A single training epoch was chosen because the 162K-sample corpus with task-weighted sampling provides sufficient coverage of all tasks without the overfitting risk of multiple passes over the data.
Maximum sequence length of 1024 tokens covers the large majority of training samples.
This configuration yields 1.6\% trainable parameters for Qwen2.5 and 1.2\% for Qwen3, with LoRA adapter sizes of approximately 240\,MB each.
Training was performed on a single NVIDIA RTX 6000 Ada GPU (48\,GB) per model, with total training time of approximately 26 hours per model.

\subsection{Evaluation Protocol}

\textbf{Data Split.}
For each task, we create a held-out test set by randomly sampling 500 instances (seed=42) that are disjoint from the training data.
For RADS assignment, the 500 test samples are drawn from the Multi-RADS dataset's held-out test split.
For MIMIC-CXR-derived tasks (impression, abnormality detection), test samples are drawn from reports not included in the training subset.
For N/M-staging, training and test samples were randomly partitioned (seed=42) from CT reports in the Merlin dataset~\cite{blankemeier2026merlin} that were screened for explicit lymph node or distant metastasis descriptors; no predefined test split from the original dataset was used.
For NLI, the full 480-sample RADNLI test set is used (no sampling needed, as it is already small).
For NER, temporal comparison, and Q\&A, test samples are randomly held out from the respective source datasets before any subsampling or capping is applied.

\textbf{Metrics.}
We evaluate all 9 tasks with task-specific metrics (Table~\ref{tab:metrics}):

\begin{itemize}[leftmargin=*]
  \item \textbf{RADS}: Validity (fraction of outputs matching a valid RADS category format) and Accuracy (exact string match of predicted and ground truth RADS category).
  \item \textbf{Impression, NER, QA}: ROUGE-L F1 score~\cite{lin2004rouge}, measuring longest common subsequence overlap between predicted and reference text.
  \item \textbf{N/M Staging, NLI}: Classification accuracy (exact match of predicted and ground truth labels).
  \item \textbf{Temporal Comparison}: Set-Jaccard overlap between predicted and ground truth sets of temporal change labels.
  \item \textbf{Abnormality Detection}: Per-label accuracy averaged across all abnormality categories.
\end{itemize}

\begin{table}[t]
\centering
\small
\caption{Evaluation metrics by task. $n$ = test set size per task.}
\label{tab:metrics}
\begin{tabular}{l l l}
\toprule
\textbf{Task} & \textbf{Metric} & \textbf{$n$} \\
\midrule
RADS Assignment & Accuracy (Acc), Validity (Val) & 500 \\
Impression Gen. & ROUGE-L & 500 \\
Temporal Comp. & Jaccard Index & 500 \\
Radiology NER & ROUGE-L & 500 \\
N-staging & Accuracy & 500 \\
Abnormality Det. & Per-label Accuracy & 500 \\
Radiology Q\&A & ROUGE-L & 500 \\
M-staging & Accuracy & 500 \\
Radiology NLI & Accuracy & 480 \\
\bottomrule
\end{tabular}
\end{table}

For Qwen3's thinking mode, we strip \texttt{<think...</think<} tags from all outputs before evaluation.
For zero-shot baselines, Qwen3's thinking is disabled via the \texttt{enable\_thinking=False} parameter in the chat template; when enabled, the model exhausts generation budgets on internal reasoning and produces no answer.

\section{Results}

\subsection{Zero-Shot vs.\ Fine-Tuned Performance}

\begin{figure}[t]
\centering
\includegraphics[width=\columnwidth]{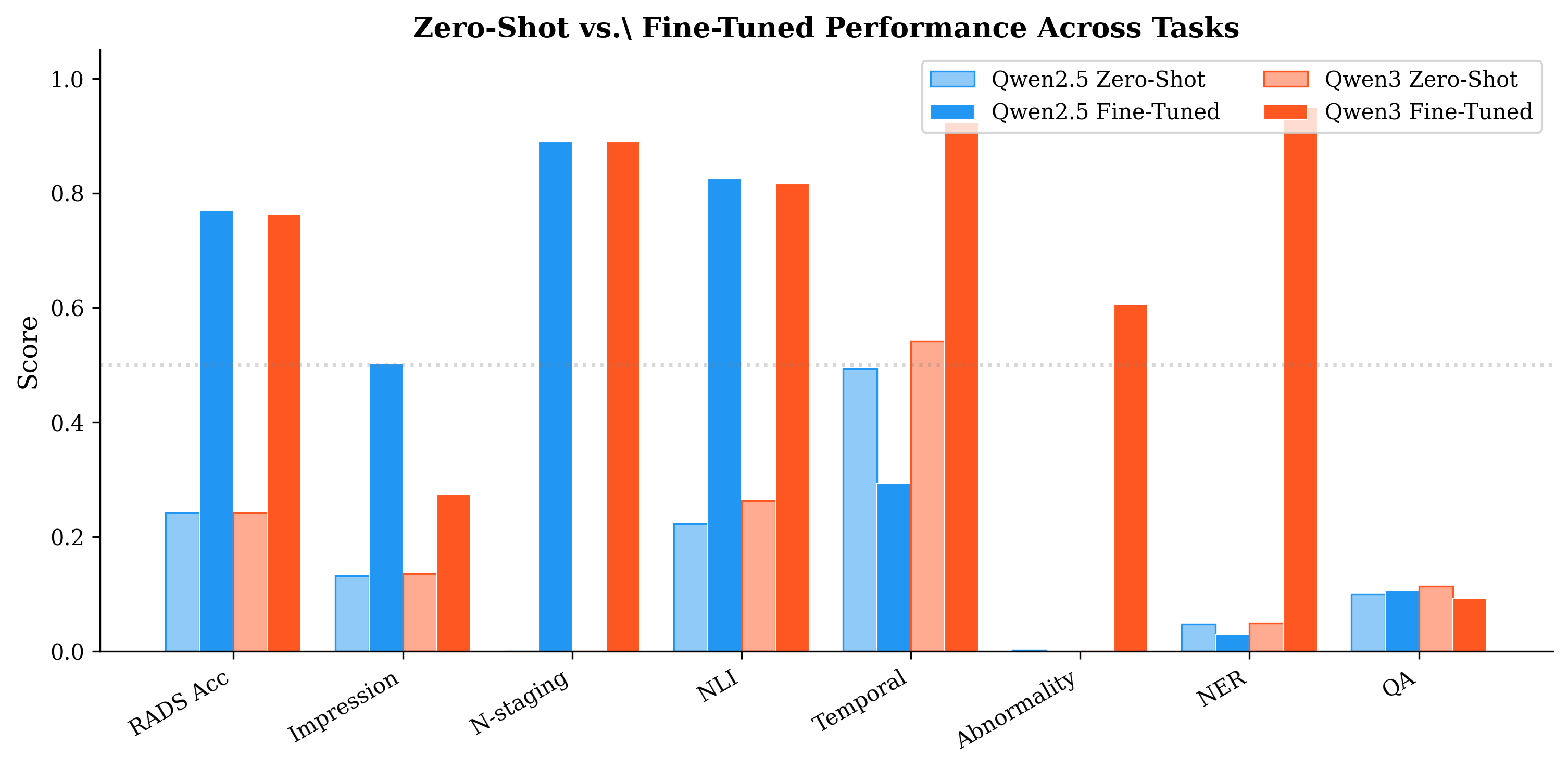}
\caption{Zero-shot vs.\ fine-tuned performance across 9 radiology tasks for Qwen2.5-3B (blue) and Qwen3-4B (orange). Bars show absolute metric values; arrows indicate direction of change from zero-shot to fine-tuned. Note the complementary pattern: Qwen2.5 excels at generation tasks while Qwen3 excels at extraction tasks.}
\label{fig:zs_vs_ft}
\end{figure}

Figure~\ref{fig:zs_vs_ft} and Table~\ref{tab:main} present the main comparison between zero-shot baselines and LoRA fine-tuned models across all 9 tasks.

\begin{table*}[htbp]
\centering
\resizebox{\textwidth}{!}{%
\begin{tabular}{l l c c c | c c c | c}
\toprule
\multirow{2}{*}{\textbf{Task}} & \multirow{2}{*}{\textbf{Metric}} & \multicolumn{3}{c|}{\textbf{Qwen2.5-3B}} & \multicolumn{3}{c|}{\textbf{Qwen3-4B}} & \textbf{Oracle} \\
\cmidrule(lr){3-5} \cmidrule(lr){6-8} \cmidrule(lr){9-9}
& & ZS & FT & $\Delta$ & ZS & FT & $\Delta$ & Best \\
\midrule
RADS Assignment & Acc & 0.242 & \textbf{0.770} & +.528 & 0.242 & 0.764 & +.522 & \textbf{0.770} \\
RADS Assignment & Val & 0.732 & 0.990 & +.258 & 0.842 & \textbf{1.000} & +.158 & \textbf{1.000} \\
Impression Gen. & RL & 0.132 & \textbf{0.502} & +.370 & 0.135 & 0.274 & +.138 & \textbf{0.502} \\
N-staging & Acc & 0.000 & \textbf{0.890} & +.890 & 0.000 & \textbf{0.890} & +.890 & 0.890 \\
M-staging & Acc & 0.000 & \textbf{0.730} & +.730 & 0.000 & \textbf{0.730} & +.730 & 0.730 \\
Radiology NLI & Acc & 0.223 & \textbf{0.825} & +.602 & 0.263 & 0.817 & +.554 & \textbf{0.825} \\
Temporal Comp. & Jac & 0.493 & 0.293 & $-$.200 & 0.542 & \textbf{0.923} & +.381 & \textbf{0.923} \\
Abnormality Det. & LA & 0.002 & 0.000 & $-$.002 & 0.000 & \textbf{0.606} & +.606 & \textbf{0.606} \\
Radiology NER & RL & 0.047 & 0.030 & $-$.017 & 0.049 & \textbf{0.950} & +.902 & \textbf{0.950} \\
Radiology QA & RL & 0.100 & \textbf{0.107} & +.007 & 0.113 & 0.093 & $-$.021 & \textbf{0.107} \\
\bottomrule
\end{tabular}%
}
\caption{Multi-task performance: zero-shot (ZS) baselines vs.\ LoRA fine-tuned (FT) models on held-out test sets ($n$=500 per task unless noted). $\Delta$ = FT$-$ZS. Best FT result per task is \textbf{bolded}. Oracle = task-routed best-of-both-models ensemble. Acc = Accuracy; Val = Validity; RL = ROUGE-L; Jac = Jaccard Index; LA = Per-label Accuracy.}
\label{tab:main}
\end{table*}

Fine-tuning produces dramatic improvements across most tasks for both models:

\begin{itemize}[leftmargin=*]
  \item \textbf{RADS accuracy} improves from 24.2\% (zero-shot for both models) to 77.0\% (Qwen2.5) and 76.4\% (Qwen3), a $+$52--53 percentage point gain.
  \item \textbf{N-staging} jumps from 0\% to 89\%, and \textbf{M-staging} from 0\% to 73\% for both models, demonstrating that structured clinical outputs require task-specific training.
  \item \textbf{NLI accuracy} more than triples: 22.3\%$\rightarrow$82.5\% (Qwen2.5), 26.3\%$\rightarrow$81.7\% (Qwen3).
  \item \textbf{Impression generation} ROUGE-L improves by 280\% for Qwen2.5 (0.132$\rightarrow$0.502) but only 103\% for Qwen3 (0.135$\rightarrow$0.274).
  \item \textbf{NER} sees a dramatic approximately 1,840\% improvement for Qwen3 (0.049$\rightarrow$0.950) but negligible change for Qwen2.5.
\end{itemize}

\begin{figure}[t]
\centering
\includegraphics[width=\columnwidth]{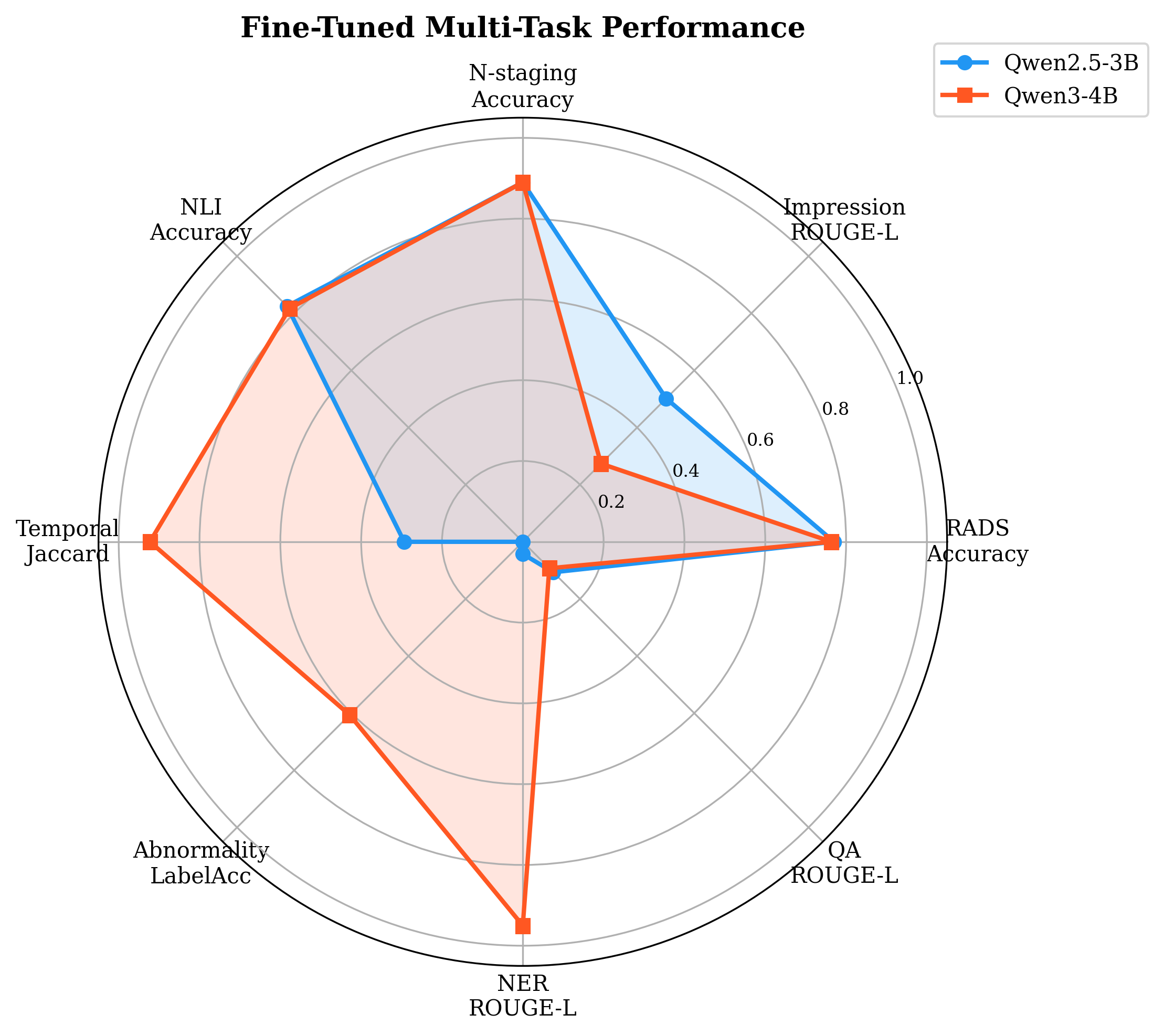}
\caption{Radar plot comparing fine-tuned Qwen2.5-3B and Qwen3-4B across 9 tasks. Each axis represents a task metric (normalized to 0--1 scale). The complementary pattern is clearly visible: Qwen2.5 (blue) dominates the upper-left quadrant (generation tasks) while Qwen3 (orange) dominates the lower-right (extraction tasks).}
\label{fig:radar}
\end{figure}

A surprising finding is that fine-tuning can \emph{degrade} zero-shot capabilities for certain model-task combinations (Figure~\ref{fig:radar}).
Qwen2.5's temporal comparison drops from 0.493 (zero-shot) to 0.293 after fine-tuning, while Qwen3's improves from 0.542 to 0.923.
This suggests that multi-task training can cause negative interference for some architectures on specific tasks, and that model architecture plays a critical role in determining which tasks benefit from fine-tuning.

\subsection{Complementary Model Strengths}

A striking finding is the complementary nature of the two models (Figure~\ref{fig:radar}).
As shown in Table~\ref{tab:main}, each model wins or ties on 5 out of 10 metrics:

\begin{itemize}[leftmargin=*]
  \item \textbf{Qwen2.5-3B excels at} structured generation: RADS classification (0.770 vs 0.764), impression generation (0.502 vs 0.274, +83\%), NLI (0.825 vs 0.817), and QA (0.107 vs 0.093).
  \item \textbf{Qwen3-4B excels at} extraction and detection: temporal comparison (0.923 vs 0.293, +215\%), abnormality detection (0.606 vs 0.000), NER (0.950 vs 0.030, +3,067\%), and RADS validity (1.000 vs 0.990).
  \item \textbf{Ties}: N-staging (both 0.890) and M-staging (both 0.730).
\end{itemize}

Notably, Qwen2.5 fine-tuning \emph{hurts} temporal and NER performance compared to zero-shot, while Qwen3 fine-tuning produces massive gains on the same tasks.
Conversely, Qwen2.5 achieves 83\% higher ROUGE-L on impression generation.
This suggests fundamental architectural differences: Qwen2.5's standard decoder excels at autoregressive text generation, while Qwen3's architecture (with improved attention mechanisms) is better suited for extractive token-level tasks.

\subsection{Per-RADS System Analysis}

\begin{figure}[t]
\centering
\includegraphics[width=\columnwidth]{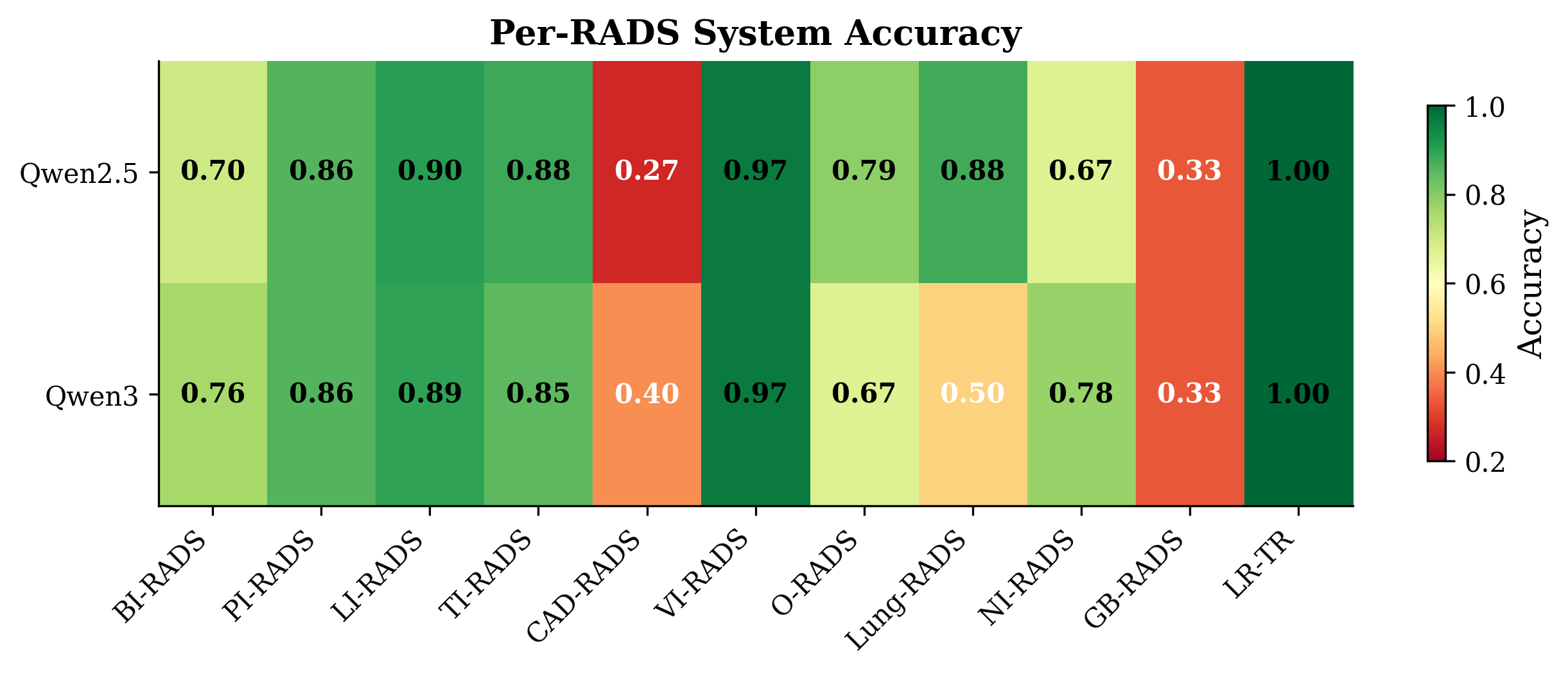}
\caption{Per-RADS system accuracy heatmap comparing Qwen2.5-3B (left) and Qwen3-4B (right). Darker green indicates higher accuracy. Both models show strong performance on common systems (VI-RADS, TI-RADS, PI-RADS) and struggle with rare systems (CAD-RADS, GB-RADS). Qwen3 shows a notable advantage on BI-RADS (the largest category), while Qwen2.5 is more consistent across systems.}
\label{fig:rads_heatmap}
\end{figure}

Figure~\ref{fig:rads_heatmap} and Table~\ref{tab:rads} break down RADS classification performance by system.

\begin{table}[t]
\centering
\small
\caption{Per-RADS system accuracy (fine-tuned, $n$=500 total). Both models evaluated on the same test set. Q2.5 = Qwen2.5-3B-Instruct; Q3 = Qwen3-4B. $^*$LI-RADS includes LR-TR treatment response sub-category (72 LI-RADS + 8 LR-TR samples).}
\label{tab:rads}
\begin{tabular}{l r c c c}
\toprule
\textbf{System} & \textbf{$n$} & \textbf{Q2.5} & \textbf{Q3} & \textbf{Best} \\
\midrule
BI-RADS   & 165 & 0.697 & \textbf{0.770} & Q3 \\
CAD-RADS  &  15 & 0.267 & \textbf{0.400} & Q3 \\
GB-RADS   &   3 & 0.333 & \textbf{0.333} & Tie \\
LI-RADS$^*$   &  80 & \textbf{0.750} & 0.662 & Q2.5 \\
Lung-RADS &   8 & \textbf{0.875} & 0.500 & Q2.5 \\
NI-RADS   &   9 & 0.667 & \textbf{0.778} & Q3 \\
O-RADS    &  33 & \textbf{0.788} & 0.667 & Q2.5 \\
PI-RADS   &  71 & \textbf{0.859} & 0.845 & Q2.5 \\
TI-RADS   &  86 & \textbf{0.884} & 0.849 & Q2.5 \\
VI-RADS   &  30 & \textbf{0.967} & 0.967 & Tie \\
\midrule
\textbf{Overall} & 500 & \textbf{0.770} & 0.764 & Q2.5 \\
\bottomrule
\end{tabular}
\end{table}

Both models achieve strong performance on common systems (VI-RADS 96.7\%, TI-RADS 85\%, PI-RADS 85\%) but struggle with rare or complex systems (CAD-RADS 27--40\%, GB-RADS 33\%).
Qwen2.5 shows more consistent performance across systems, while Qwen3 excels specifically on BI-RADS - the most clinically prevalent system.
Performance generally correlates with training data availability: systems with more training samples (BI-RADS: 3,124, TI-RADS: 1,684) achieve higher accuracy than those with few samples (GB-RADS: 77, Lung-RADS: 121).

\subsection{Few-Shot Prompting Hurts Fine-Tuned Models}

\begin{figure}[t]
\centering
\includegraphics[width=\columnwidth]{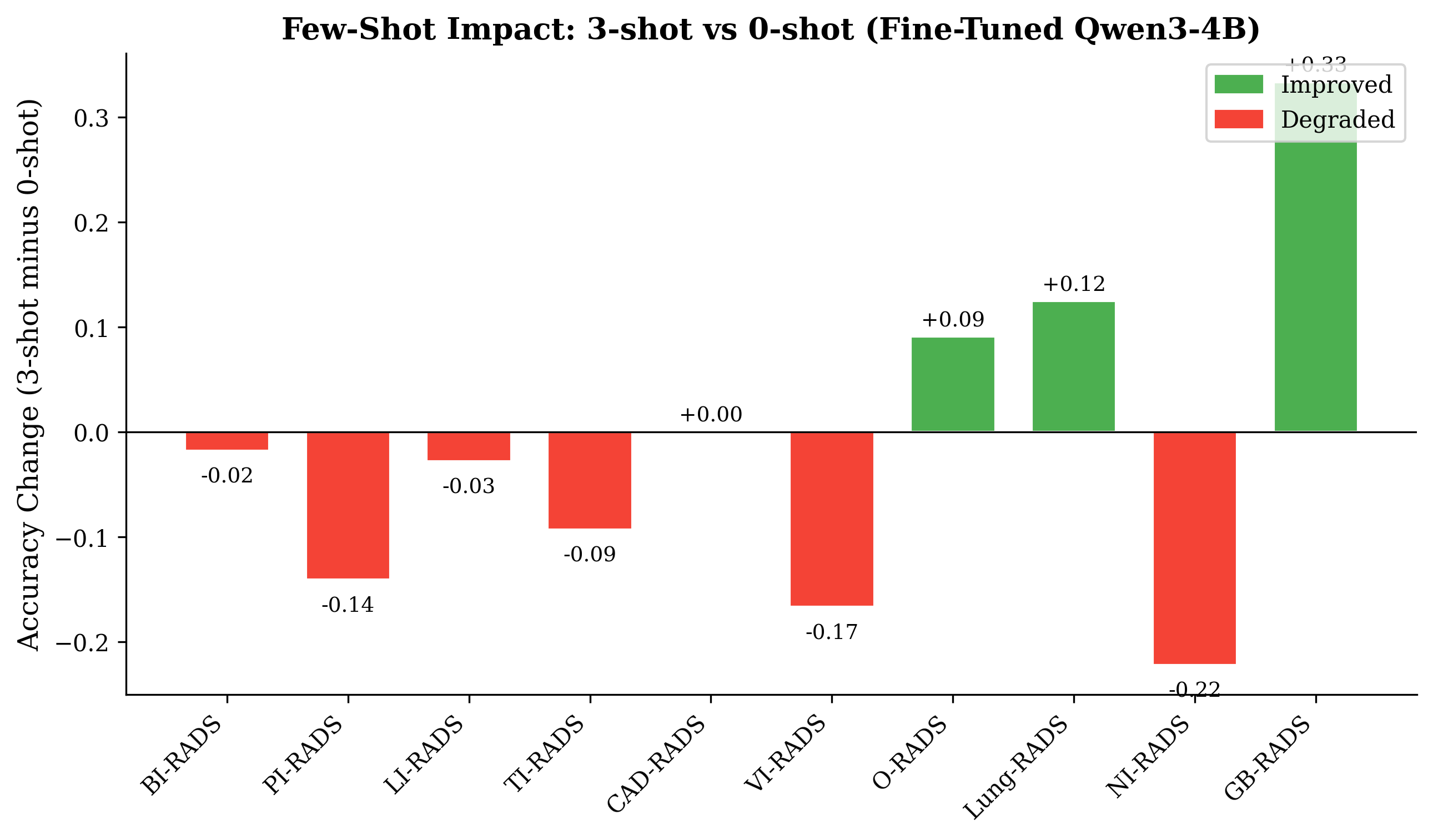}
\caption{Impact of few-shot prompting on fine-tuned Qwen3-4B RADS accuracy. Bars show per-system accuracy under zero-shot (blue) and 3-shot (orange) conditions. Few-shot prompting helps rare systems (GB-RADS, Lung-RADS) but hurts well-learned systems (NI-RADS, VI-RADS, PI-RADS), resulting in a net decrease of 5.0 percentage points.}
\label{fig:fewshot}
\end{figure}

We evaluate 3-shot prompting with the fine-tuned Qwen3-4B model on RADS assignment, selecting same-system examples from the training set.
Figure~\ref{fig:fewshot} and Table~\ref{tab:fewshot} show the results.

\begin{table}[t]
\centering
\small
\caption{Few-shot prompting with fine-tuned Qwen3-4B on RADS assignment ($n$=500). Val = Validity; Acc = Accuracy.}
\label{tab:fewshot}
\begin{tabular}{l c c}
\toprule
\textbf{Setting} & \textbf{Val} & \textbf{Acc} \\
\midrule
Zero-shot (fine-tuned) & 1.000 & 0.764 \\
3-shot (fine-tuned)    & 1.000 & 0.714 \\
\midrule
\multicolumn{3}{l}{\textit{Per-system changes (0-shot $\rightarrow$ 3-shot):}} \\
\quad GB-RADS ($n$=3)    & 0.333 $\rightarrow$ 0.667 & +33.3\% \\
\quad Lung-RADS ($n$=8)  & 0.500 $\rightarrow$ 0.625 & +12.5\% \\
\quad O-RADS ($n$=33)    & 0.667 $\rightarrow$ 0.758 & +9.1\% \\
\quad PI-RADS ($n$=71)   & 0.845 $\rightarrow$ 0.704 & $-$14.1\% \\
\quad VI-RADS ($n$=30)   & 0.967 $\rightarrow$ 0.800 & $-$16.7\% \\
\quad NI-RADS ($n$=9)    & 0.778 $\rightarrow$ 0.556 & $-$22.2\% \\
\bottomrule
\end{tabular}
\end{table}

Few-shot prompting \emph{reduces} overall accuracy by 5.0 percentage points (0.764 $\rightarrow$ 0.714).
While it helps rare systems with limited training data (GB-RADS +33\%, Lung-RADS +12.5\%), it \emph{hurts} systems where the model already performs well (NI-RADS $-$22\%, VI-RADS $-$17\%, PI-RADS $-$14\%).
This is consistent with the \textit{distribution shift} hypothesis~\cite{zhao2021calibrate}: fine-tuned models learn to map specific prompt formats to outputs.
Adding in-context examples changes the input distribution, degrading the learned mapping.
This finding has practical implications - for domain-specialized SLMs, fine-tuning is more effective than in-context learning, and combining both strategies can be counterproductive.

\subsection{Error Analysis and Clinical Safety}

\begin{figure}[t]
\centering
\includegraphics[width=\columnwidth]{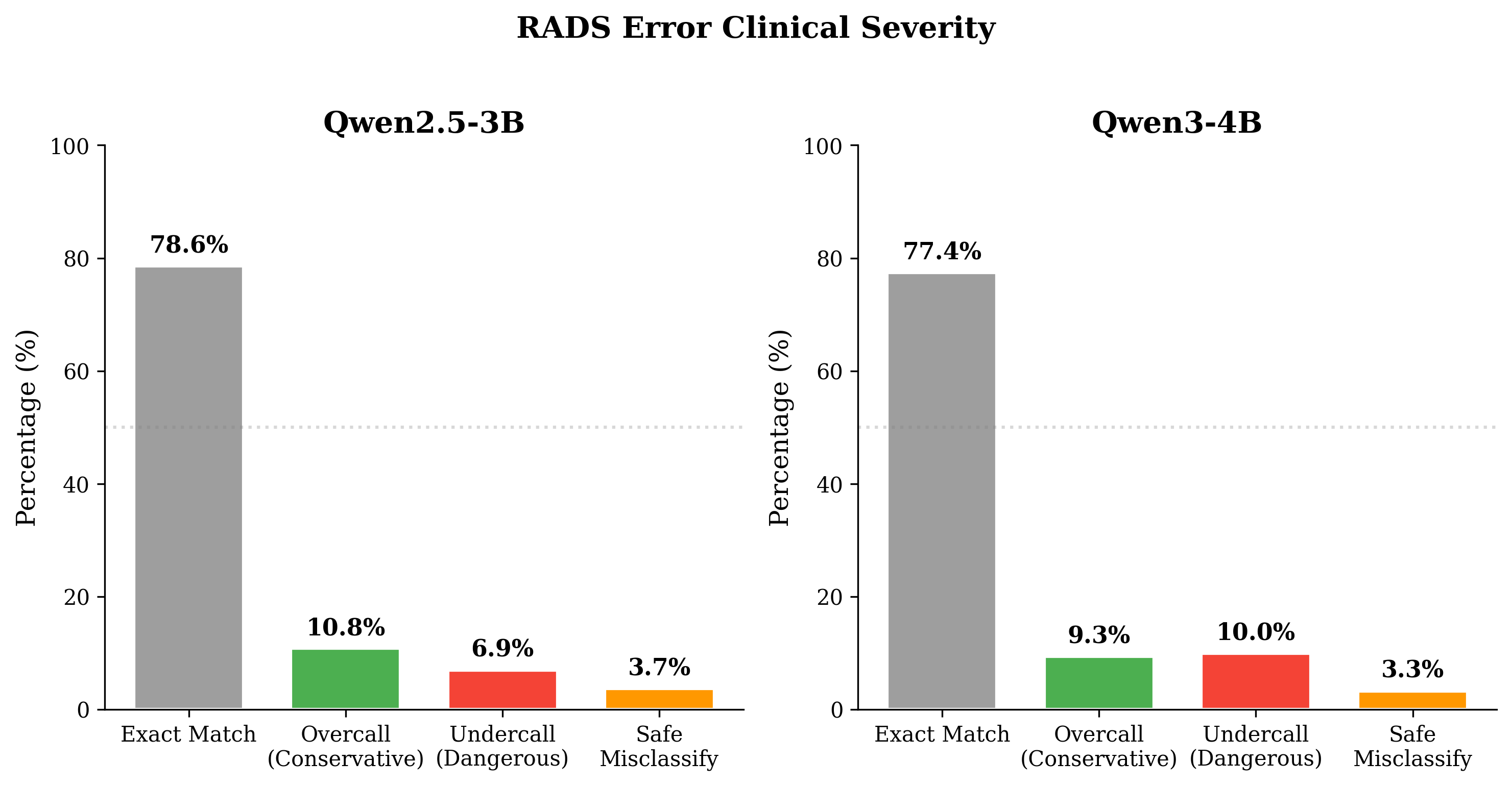}
\caption{Clinical severity analysis of RADS classification errors. Left: error direction distribution for Qwen2.5-3B (top) and Qwen3-4B (bottom). Right: per-model undercall vs.\ overcall ratio among incorrect predictions. Qwen2.5 errs conservatively (50.5\% overcalls), making it clinically safer than Qwen3 (44.1\% undercalls).}
\label{fig:severity}
\end{figure}

Figure~\ref{fig:severity} and Table~\ref{tab:severity} present a clinical severity analysis of RADS classification errors.

\begin{table}[t]
\centering
\small
\caption{Clinical severity of RADS classification errors$^\dagger$ ($n$=500). Undercall = predicted less severe than ground truth (dangerous); Overcall = predicted more severe (conservative); Safe = same severity level, different system. Q2.5 = Qwen2.5-3B-Instruct; Q3 = Qwen3-4B. $^\dagger$Remaining cases (Qwen2.5: 10; Qwen3: 8) represent cross-system predictions where undercall/overcall direction is undefined.}
\label{tab:severity}
\begin{tabular}{l c c}
\toprule
\textbf{Error Type} & \textbf{Q2.5} & \textbf{Q3} \\
\midrule
Exact match & 385 (78.6\%) & 381 (77.4\%) \\
\textbf{Undercall (dangerous)} & \textbf{34 (6.9\%)} & \textbf{49 (10.0\%)} \\
Overcall (conservative) & 53 (10.8\%) & 46 (9.3\%) \\
Safe misclassify & 18 (3.7\%) & 16 (3.3\%) \\
\midrule
\multicolumn{3}{l}{\textit{Error distribution (of incorrect predictions):}} \\
\quad Undercall ratio & 32.4\% & 44.1\% \\
\quad Overcall ratio & 50.5\% & 41.4\% \\
\quad Safe ratio & 17.1\% & 14.4\% \\
\bottomrule
\end{tabular}
\end{table}

A critical finding for clinical deployment is the error \emph{direction}.
Qwen2.5 errs conservatively: 50.5\% of its errors are overcalls (predicting more severe than ground truth), while only 32.4\% are dangerous undercalls.
This conservative bias is clinically preferable - overcalling leads to additional follow-up imaging, while undercalling can miss pathology~\cite{acrbirads2013}.
Qwen3 shows a more balanced but riskier error profile with 44.1\% undercalls vs 41.4\% overcalls.
Overall, Qwen2.5 has 3.0 percentage points fewer dangerous undercalls (6.9\% vs 10.0\%).

Most RADS errors are ``off-by-one'' (adjacent category confusion) rather than gross misclassification.
Across both models, 62--76\% of errors involve adjacent categories (e.g., BI-RADS 3 predicted as BI-RADS 4), while only 4--14\% are off-by-two-or-more, suggesting the models capture the ordinal structure of RADS scales.

McNemar's test confirms that the two models are not significantly different on RADS accuracy ($p$=0.73) or NLI accuracy ($p$=0.62), with overlapping 95\% bootstrap confidence intervals on both tasks (Table~\ref{tab:stats}).
N-staging and M-staging also show no difference ($p$=1.0), as both models produce identical predictions on these tasks.
However, Wilcoxon signed-rank tests reveal highly significant differences ($p < 0.001$) on the five remaining tasks, confirming the complementary strengths: Qwen2.5 significantly outperforms Qwen3 on impression generation and QA, while Qwen3 significantly outperforms Qwen2.5 on temporal comparison, abnormality detection, and NER.

\begin{table}[t]
\centering
\footnotesize
\caption{Statistical significance tests across all 9 tasks. Classification tasks use McNemar's test; generation/extraction tasks use Wilcoxon signed-rank test on per-sample scores. ns = not significant; ***$p < 0.001$.}
\label{tab:stats}
\begin{tabular}{l c c c c}
\toprule
\textbf{Task} & \textbf{Qwen2.5 FT} & \textbf{Qwen3 FT} & \textbf{Test} & \textbf{$p$-value} \\
\midrule
RADS Accuracy   & 0.770 & 0.764 & McNemar  & 0.727 (ns) \\
Radiology NLI   & 0.825 & 0.817 & McNemar  & 0.620 (ns) \\
N-staging       & 0.890 & 0.890 & McNemar  & 1.000 (ns) \\
M-staging       & 0.730 & 0.730 & McNemar  & 1.000 (ns) \\
\midrule
Impression Gen. & 0.502 & 0.274 & Wilcoxon & $<$0.001 (***) \\
Radiology NER   & 0.030 & 0.950 & Wilcoxon & $<$0.001 (***) \\
Radiology QA    & 0.107 & 0.093 & Wilcoxon & $<$0.001 (***) \\
Temporal Comp.  & 0.293 & 0.923 & Wilcoxon & $<$0.001 (***) \\
Abnormality Det.& 0.000 & 0.606 & Wilcoxon & $<$0.001 (***) \\
\bottomrule
\end{tabular}
\end{table}

\subsection{NLI Error Patterns}

Both models show a consistent confusion pattern on the NLI task (Table~\ref{tab:nli_confusion}).
The ``neutral'' class dominates the dataset (58.5\% of samples) and both models over-predict it.
The primary error mode is misclassifying entailment and contradiction as neutral (21--28 samples each), rather than confusing entailment with contradiction directly (only 2--4 samples).
This suggests the models can distinguish positive from negative relationships but struggle with the neutral ``unrelated'' category---a known challenge in NLI benchmarks~\cite{poliak2018hypothesis}.

\begin{table}[t]
\centering
\small
\caption{NLI confusion matrix for both models ($n$=480). Numbers show Qwen2.5 / Qwen3 counts. GT = Ground Truth; Entail. = Entailment; Contra. = Contradiction.}
\label{tab:nli_confusion}
\begin{tabular}{l c c c}
\toprule
\textbf{GT $\downarrow$ / Pred $\rightarrow$} & \textbf{Entail.} & \textbf{Contra.} & \textbf{Neutral} \\
\midrule
Entailment ($n$=93) & 70/73 & 2/1 & 21/19 \\
Contradiction ($n$=106) & 2/2 & 76/80 & 28/24 \\
Neutral ($n$=281) & 14/28 & 17/15 & 250/238 \\
\bottomrule
\end{tabular}
\end{table}

\subsection{Qwen3 Thinking Mode and Fine-Tuning}

An important observation concerns Qwen3's default thinking mode.
In zero-shot evaluation, Qwen3 generates extensive internal reasoning (\texttt{<think...>} tags) before producing outputs.
With limited generation budgets (e.g., 30 tokens for RADS), the model exhausts its budget on reasoning and fails to produce structured answers, achieving only 16\% RADS validity even with 512 additional thinking tokens.
Critically, LoRA fine-tuning \emph{implicitly suppresses} this thinking behavior.
The fine-tuned Qwen3 model directly produces structured outputs without reasoning, because the training data contains no thinking tokens.
This demonstrates that fine-tuning not only teaches domain knowledge but also shapes the model's response strategy - an important consideration when deploying thinking-capable models in structured output scenarios.

\subsection{CPU Deployment Benchmarks}

\begin{figure}[t]
\centering
\includegraphics[width=\columnwidth]{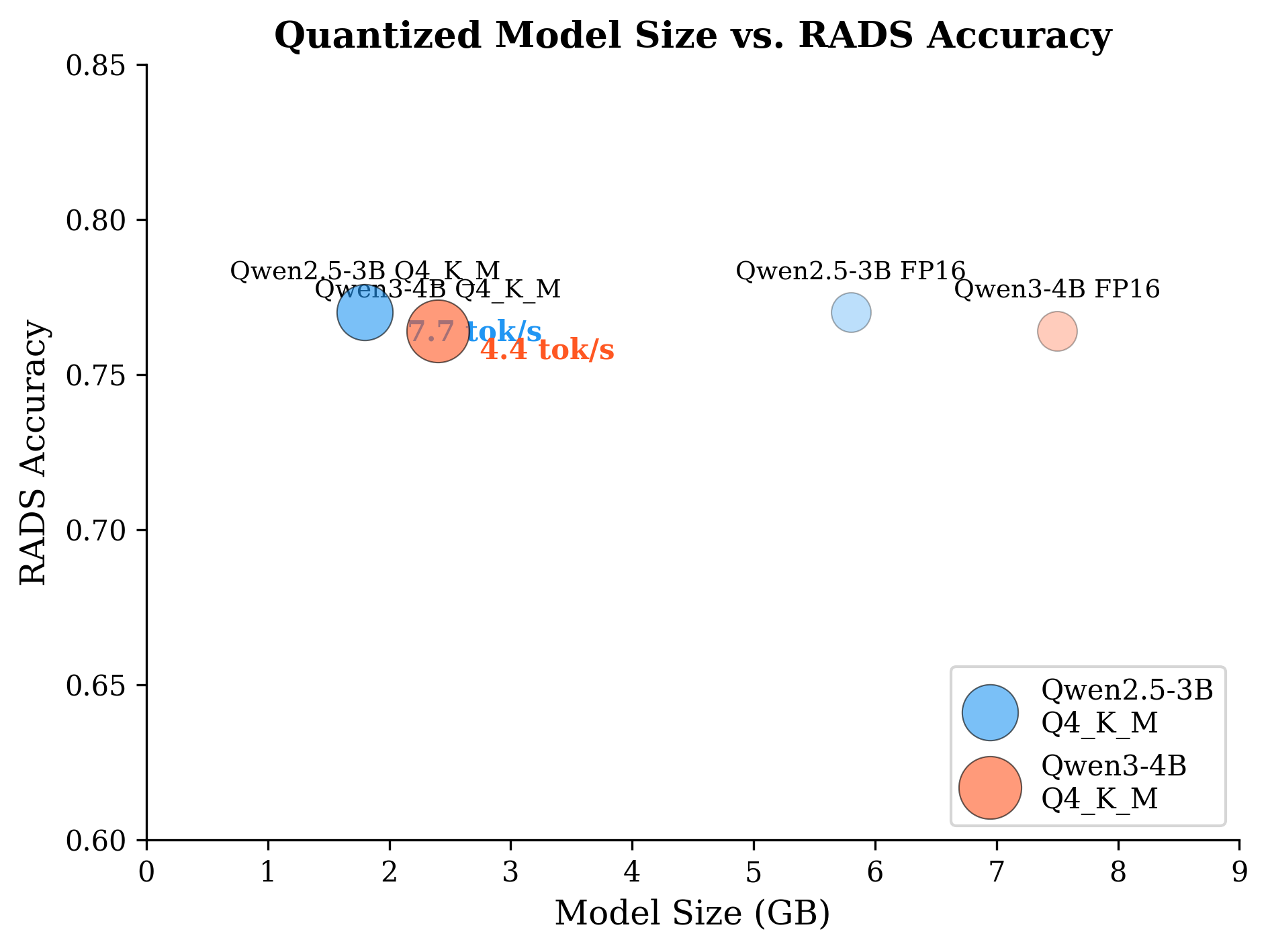}
\caption{Deployment tradeoff: model size (GGUF Q4\_K\_M) vs.\ inference throughput. RadLite models (blue markers) achieve strong multi-task performance in $\leq$2.4 GB, enabling CPU deployment. The dashed line represents typical consumer RAM constraints. Both models operate well below this threshold, with Qwen2.5-3B offering the best speed--size tradeoff at 1.8 GB and 7.7 tokens/second.}
\label{fig:deployment}
\end{figure}

Both models were quantized to GGUF format (Q4\_K\_M) and benchmarked on a consumer CPU (Intel i7-class, 4 threads).
Table~\ref{tab:gguf} and Figure~\ref{fig:deployment} summarize model sizes and inference throughput.

\begin{table}[t]
\centering
\small
\caption{GGUF quantized model sizes and CPU inference benchmarks (4 threads, Q4\_K\_M quantization, Intel i7-class CPU). tok/s = tokens per second.}
\label{tab:gguf}
\begin{tabular}{l c c c c}
\toprule
\textbf{Model} & \textbf{FP16} & \textbf{Q4\_K\_M} & \textbf{Latency} & \textbf{tok/s} \\
\midrule
Qwen2.5-3B & 5.8 GB & 1.8 GB & 2.6 s & 7.7 \\
Qwen3-4B & 7.5 GB & 2.4 GB & 4.6 s & 4.4 \\
\bottomrule
\end{tabular}
\end{table}

Both models fit comfortably in consumer RAM ($\leq$2.4 GB).
Qwen2.5-3B achieves near-real-time inference at 7.7 tokens/second, and a typical RADS query completes in approximately 2.6 seconds.
The task-routed ensemble (both models loaded simultaneously) requires only 4.2 GB total, well within consumer hardware capabilities.

\section{Discussion}

RadLite demonstrates substantial gains across all 9 radiology tasks after LoRA fine-tuning.
RADS classification accuracy reaches 77.0\% (+52 pp over zero-shot), N-staging 89\% (+89 pp), and NLI 82.5--83\% (+60 pp).
Impression generation ROUGE-L improves by 280\% for Qwen2.5 (0.132$\rightarrow$0.502), and NER ROUGE-L by approximately 1,840\% for Qwen3 (0.049$\rightarrow$0.950).
Both models can be deployed on consumer hardware at 1.8--2.4 GB (GGUF Q4\_K\_M) and operate at 4--8 tokens/second on CPU without GPU requirements, making them viable for resource-constrained clinical environments.

The Multi-RADS benchmark~\cite{multirads2025} provides relevant context for situating our RADS results.
That study evaluated 41 open-weight SLMs alongside GPT-5.2 on 10 RADS systems using 1,600 synthetic radiologist-verified reports.
GPT-5.2 achieved 81.1\% accuracy under guided prompting, open-weight models in the 20--32B range reached 73--78\%, the 1--10B class averaged 57.5\%, and sub-1B models scored only 27\%.
Direct numerical comparison with our results is not possible: the test sets differ in size, construction, and prompting regime, and our models were trained on Multi-RADS-derived data.
What is meaningful is that our zero-shot baseline of 24.2\% aligns well with the expected performance of un-adapted 3--4B models, while our fine-tuned accuracy of 77.0\% substantially bridges the gap to frontier models - consistent with the broader finding that domain-specific fine-tuning of small models can approach zero-shot performance of much larger systems~\cite{chen2023meditron,thirunavukarasu2023}.

The two models exhibit a striking complementary pattern of strengths that likely reflects fundamental architectural differences.
Qwen2.5 outperforms Qwen3 on impression generation (ROUGE-L 0.502 vs 0.274, +83\%), RADS accuracy (77.0\% vs 76.4\%), NLI (82.5\% vs 81.7\%), and QA.
Conversely, Qwen3 dramatically outperforms Qwen2.5 on NER (ROUGE-L 0.950 vs 0.030) and temporal comparison (Jaccard 0.923 vs 0.293).
Notably, Qwen2.5 fine-tuning \emph{degrades} temporal comparison and NER relative to its own zero-shot baseline, while Qwen3 fine-tuning produces massive gains on the same tasks.
We hypothesize that Qwen2.5's standard decoder architecture excels at autoregressive generation tasks, while Qwen3's enhanced architecture (with refined GQA and a pre-training mix that may emphasize structured data) is better suited for extractive token-level tasks.
This finding aligns with evidence that model architecture, not just scale, determines task-specific performance~\cite{abdin2024phi3}.
Statistical testing confirms this complementary pattern: 5 of 9 tasks show highly significant between-model differences ($p < 0.001$, Wilcoxon signed-rank), while 4 tasks (RADS, NLI, N/M-staging) show no significant difference.

A notable negative result is that 3-shot prompting with the fine-tuned Qwen3 model reduces overall RADS accuracy by 5 percentage points (0.764$\rightarrow$0.714).
Few-shot examples help data-limited systems (GB-RADS +33\%, Lung-RADS +12.5\%) but hurt well-learned systems (NI-RADS $-$22\%, VI-RADS $-$17\%, PI-RADS $-$14\%).
This is consistent with the distribution shift hypothesis~\cite{zhao2021calibrate}: fine-tuned models learn to map specific prompt formats to outputs, and adding in-context examples changes the input distribution, degrading the learned mapping.
The practical implication is clear - for domain-specialized SLMs, LoRA fine-tuning should be preferred over in-context learning.

A second negative finding is negative transfer in multi-task training.
Qwen2.5's temporal comparison degrades from 0.493 (zero-shot) to 0.293 after fine-tuning, and NER from 0.047 to 0.030.
This occurs because the shared LoRA parameters are optimized for the aggregate loss across all tasks, causing interference on tasks where the model's zero-shot representations are already well-suited.
A task-routed ensemble that dispatches generation tasks to Qwen2.5 and extraction tasks to Qwen3 circumvents this problem and achieves the best performance across all tasks.
Future work could explore LoRA composition~\cite{huang2024lorahub} or mixture-of-experts approaches to mitigate task interference while retaining the benefits of joint training.

Both models are suitable for practical clinical deployment.
Qwen2.5-3B operates at 7.7 tokens/second (1.8 GB), Qwen3-4B at 4.4 tokens/second (2.4 GB), and a RADS classification query with Qwen2.5 completes in approximately 2.6 seconds - acceptable for clinical workflow integration where a radiologist reviews the model's suggestion before confirming.
The task-routed ensemble requires only 4.2 GB combined, well within consumer hardware capabilities.
From a clinical safety perspective, Qwen2.5's conservative error profile - 50.5\% overcalls vs 32.4\% dangerous undercalls among incorrect RADS predictions - makes it preferable for screening applications where overcalling leads to additional follow-up imaging rather than missed pathology~\cite{acrbirads2013}.
Qwen3 shows a riskier profile (44.1\% undercalls), making Qwen2.5 the safer default for RADS-based clinical decision support.

Our study has several limitations.
Evaluation is limited to up to 500 samples per task, which may not capture rare edge cases or the full diversity of clinical presentations; GB-RADS ($n$=3) results are particularly statistically fragile.
The zero-shot Qwen3 baseline was hampered by thinking-mode interference, and our results likely underestimate its true zero-shot capability.
Training data is primarily English, limiting cross-lingual generalization.
We do not evaluate larger models (7B, 14B) to establish the full Pareto frontier of size vs.\ performance.
CPU benchmarks were conducted on a single architecture; performance will vary across consumer hardware.
Finally, prospective real-world validation in clinical workflows is needed before deployment.

\section{Conclusion}

We present \textbf{RadLite}, a demonstration that small language models (3--4B parameters) can achieve strong multi-task radiology performance through LoRA fine-tuning on 162K samples spanning 9 clinical tasks compiled from 12 public datasets.
Fine-tuning improves RADS classification accuracy by 53 percentage points over zero-shot baselines, and enables previously impossible tasks (N/M staging from 0\% to 89\%).
The complementary strengths of Qwen2.5-3B and Qwen3-4B suggest that a task-routed ensemble approach maximizes performance.
Our finding that few-shot prompting \emph{hurts} fine-tuned models has practical implications: for specialized SLMs, LoRA fine-tuning should be preferred over in-context learning.
All models can be quantized to $\sim$1.8--2.4GB for CPU deployment at 4--8 tokens/second, making them practical for clinical use on consumer hardware without GPU requirements.
This work establishes a foundation for deploying multi-task radiology AI in resource-constrained clinical environments.
Code, quantized models, and inference notebooks are available at \url{https://github.com/RadioX-Labs/RadLite}.

\end{document}